\pdfoutput=1

\documentclass[11pt]{article}

\usepackage{ACL2023}

\usepackage{times}
\usepackage{latexsym}

\usepackage[T1]{fontenc}

\usepackage[utf8]{inputenc}

\usepackage{microtype}

\usepackage{inconsolata}

\usepackage{paralist}
\usepackage{adjustbox}

\usepackage{amsmath,amssymb,amsfonts}
\usepackage{pifont}

\usepackage{multirow}
\usepackage{booktabs}
\usepackage{arydshln}
\usepackage{bm}
\usepackage{color,xcolor}
\usepackage{colortbl}
\usepackage{soul}
\usepackage{utfsym}

\definecolor{highlightgreen1}{RGB}{229,250,245}
\definecolor{highlightgreen2}{RGB}{0,120,87}
\definecolor{purple1}{RGB}{175,20,241}
\definecolor{green1}{RGB}{0,176,80}

\definecolor{ired}{RGB}{196,44,46}
\definecolor{igreen}{RGB}{5,209,75}

\definecolor{nmgray}{RGB}{229,229,229}
\definecolor{underlinegray}{RGB}{197,197,197}

\setul{0.3ex}{0.2ex}
\setulcolor{underlinegray}

\title{\textsc{Cross$^2$StrA}: Unpaired Cross-lingual Image Captioning\\ with \ul{Cross}-lingual \ul{Cross}-modal \ul{Str}ucture-pivoted \ul{A}lignment}

\author{
Shengqiong Wu,  \quad
Hao Fei\Thanks{ Corresponding author: Hao Fei}, \quad
Wei Ji,  \quad
Tat-Seng Chua \\
Sea-NExT Joint Lab, School of Computing, National University of Singapore \\
\tt { swu@u.nus.edu \quad \{haofei37, jiwei, dcscts\}@nus.edu.sg,}
\\
}

\begin{document}
\maketitle
\begin{abstract}
Unpaired cross-lingual image captioning has long suffered from irrelevancy and disfluency issues, due to the inconsistencies of the semantic scene and syntax attributes during transfer.
In this work, we propose to address the above problems by incorporating the scene graph (SG) structures and the syntactic constituency (SC) trees.
Our captioner contains the \emph{semantic structure-guided image-to-pivot captioning} and the \emph{syntactic structure-guided pivot-to-target translation}, two of which are joined via pivot language.
We then take the SG and SC structures as pivoting, performing cross-modal semantic structure alignment and cross-lingual syntactic structure alignment learning.
We further introduce cross-lingual\&cross-modal back-translation training to fully align the captioning and translation stages.
Experiments on English$\leftrightarrow$Chinese transfers show that our model shows great superiority in improving captioning relevancy and fluency.
\end{abstract}


\section{Introduction}

Generating texts to describe images (a.k.a., image captioning) has many real-world applications, such as virtual assistants and image indexing \cite{FangGISDDGHMPZZ15}.
Current image captioning models have achieved impressive performance \cite{JiaGFT15,GuC0C18,JiLSCLW0J21}, yet are mainly limited to the English language due to the large-scale paired image-caption datasets.
Subject to the scarcity of paired captioning data, the development of captioning in other languages is thus greatly hindered.
While manually crafting sufficient paired data is prohibitively expensive, cross-lingual image captioning \cite{miyazaki-shimizu-2016-cross} offers a promising solution, which aims to transfer a captioner trained at resource-rich language (e.g., English) to the resource-scarce language(s) without paired captioning data at target language(s).

\begin{figure*}[!t]
\centering
\includegraphics[width=1\textwidth]{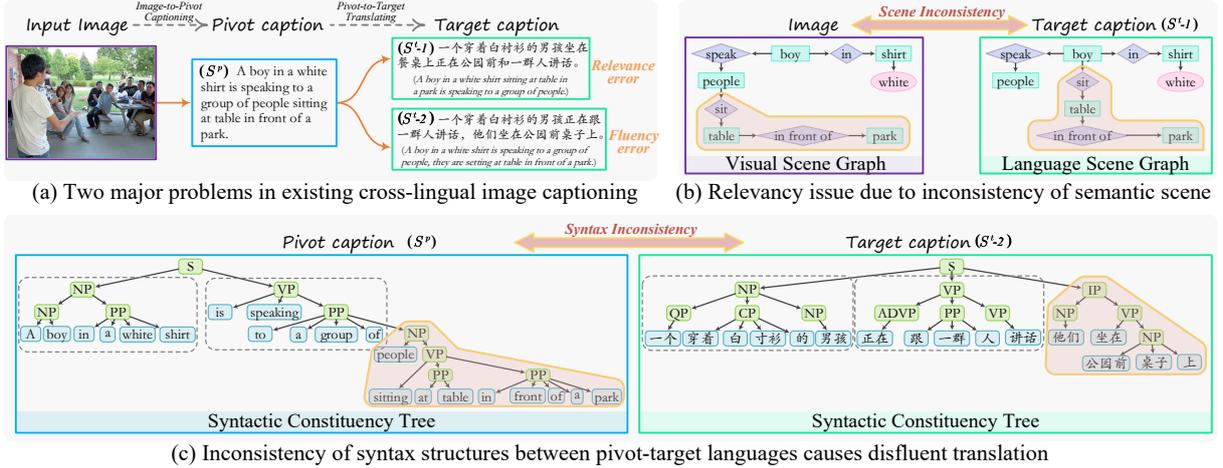}
\caption{
We illustrate two issues in translation-based cross-lingual captioning (a), caused by the inconsistencies of semantic scene structures (b) and syntax structures (c).
We simplify the constituency trees for better understanding, and the dotted grey box areas indicate the counterpart constituents across different languages.
(Best viewed in color)
}
\label{intro}
\end{figure*}

A direct approach is to make use of the current translation techniques, i.e., the pivot language translation method.
Here pivot language is the resource-rich language, e.g., English.
For example, the pivot-side captioner first generates pivot captions for images, which are then translated into the target-side captions.
Or one can create the pseudo image-caption pairs for directly training a target-side captioner, by translating the pivot training captions into the target ones \cite{LanLD17}.
However, the above translation-based method suffers from two major issues (cf. \ref{intro}(a)), including \emph{irrelevancy} and \emph{disfluency} \cite{0003CZJ19}.
On the one hand, due to the lack of paired vision contexts, a translated description can easily deviate from the original visual semantics, leading to ambiguous or inaccurate captioning.
On the other hand, restricted to the translation system itself, translated texts often suffer from disfluent language, especially for the lengthy and complex descriptions.

Some previous efforts are carried out to rectify the above two key errors for better cross-lingual captioning.
\citet{LanLD17} solve the translation disfluency issue by estimating the fluency of translation texts, then rejecting those disfluent ones.
Yet their method dramatically sacrifices the paired training data, and meanwhile suffers from low-efficiency owing to the incremental screening process.
\citet{0003CZJ19} propose to enhance the relevance and fluency of translations by designing some rewards via the reinforcement learning technique.
However, the \emph{REINFORCE} algorithm \cite{Williams92} is hard to train, and easily leads to unstable results.
We note that there are two critical abilities a cross-lingual captioning system should possess to solve the corresponding problems.
For content relevancy, the kernel lies in sufficiently modeling the vision-language semantic alignment; while for language fluency, it is key to effectively capture the gaps of linguistic attributes and characteristics between the pivot and target languages.

Besides the translation-based methods, the pivoting-based cross-lingual captioning methods have shown effectiveness, where the whole task learning is broken down into two steps, image-to-pivot captioning and pivot-to-target translation \cite{GuJCW18,GaoZYJG22}.
The image-to-pivot captioning learns to describe images in the pivot language based on pivot-side paired captioning data, and the pivot-to-target translation is performed based on parallel sentences.
Two cross-model and cross-lingual subtasks are trained on two separate datasets, and aligned by the pivot language.
Although achieving improved task performances, existing pivoting-based methods \cite{GuJCW18,GaoZYJG22} still fail to fully address the two major problems of cross-lingual captioning, due to the insufficient alignment of either vision-language semantics or pivot-target syntax.

\begin{figure}[!t]
\centering
\includegraphics[width=0.99\columnwidth]{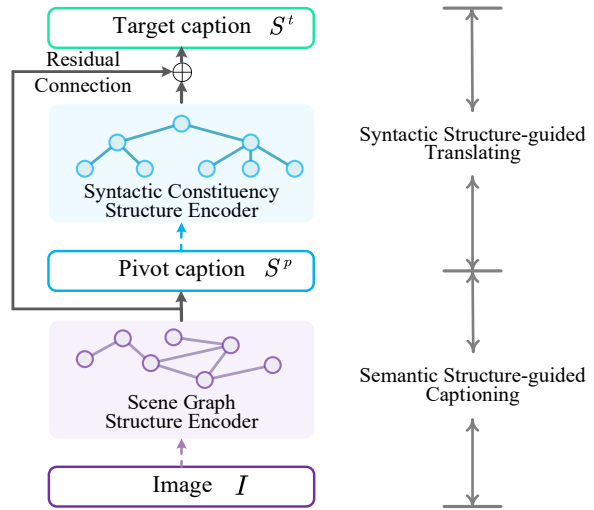}
\caption{
A high-level schematic of the cross-lingual captioning model architecture.
}
\label{framework}
\vspace{-3mm}
\end{figure}

\vspace{-1mm}
To this end, we present a novel syntactic and semantic structure-guided model for cross-lingual image captioning.
We build the framework based on the pivoting-based scheme, as shown in Fig. \ref{framework}.
For image-to-pivot captioning, we consider leveraging the scene graphs (SG) for better image-text alignment.
Intuitively, an SG \cite{JohnsonKSLSBL15,YangTZC19} depicts the intrinsic semantic structures of texts or images, which can ideally bridge the gaps between modalities.
For the pivot-to-target translating, we make use of the syntactic constituency (SC) tree structures for better pivot-target language alignment.
Syntax features have been shown as effective supervisions for enhancing the translation quality, e.g., fluency and grammar-correctness \cite{schwartz-etal-2011-incremental,xu-etal-2020-learning,li-etal-2021-unsupervised-neural}.

\vspace{-1mm}
Based on the above framework, we further perform cross-lingual cross-modal structure-pivoted alignment learning.
First of all, we introduce an SG-pivoted cross-modal semantic structure alignment.
Based on contrastive learning \cite{LogeswaranL18,yan-etal-2021-consert} we realize the unsupervised vision-language semantic structure alignment, relieving the scene inconsistency and thus enhancing the relevancy.
Similarly, an unsupervised SC-based cross-lingual syntax structure aligning is used to learn the shared grammar transformation and thus mitigate the language disfluency during translation.
Finally, we perform the cross-lingual cross-modal back-translation training, fully aligning the two phrases of image-to-pivot captioning and pivot-to-target translation.


On English$\to$Chinese and Chinese$\to$English transfers of unpaired cross-lingual image captioning, our method achieves significant improvement over the existing best-performing methods.
Further in-depth analyses demonstrate that the integration of both scene graph and syntactic structure features is complementarily helpful in improving the captioning relevancy and disfluency of the transfer.

Our main contributions are two-fold:

$\bullet$ First, we for the first time enhance the cross-lingual image captioning by leveraging both the semantic scene graph and the syntactic constituent structure information, such that we effectively address the problems of content irrelevancy and language disfluency.

$\bullet$ Second, we propose several cross-lingual cross-modal structure-pivoted alignment learning strategies, via which we achieve effective cross-modal vision-language semantic alignment and cross-lingual pivot-target syntactic alignment.

\vspace{-1mm}
\section{Related Work}

Image captioning has been an emerging task in the past few years and received great research attention \cite{YouJWFL16,VinyalsTBE17,CorniaSBC20}.
Later, the task of cross-lingual image captioning \cite{miyazaki-shimizu-2016-cross,0003CZJ19} has been presented, to transfer the knowledge from resource-rich language to resource-poor language\footnote{Without using target-side image-caption pairs, the task is also known as unpaired cross-lingual image captioning.}, so as to spare the burden of manual data annotation for the minority languages.
However, the task has been hindered and received limited attention due to two key issues: irrelevancy and disfluency of captions. 
There are two categories of cross-lingual captioning approaches: the translation-based \cite{LanLD17,GuJCW18} and the pivoting-based \cite{GuJCW18,GaoZYJG22} methods.
The former employs an off-the-shelf translator to translate the source (pivot) captions into the target language for target-side training or as the target-side captions.
The latter reduces the noise introduction of the pipeline by jointly performing the image-to-pivot captioning step and pivot-to-target translation step, thus being the current SoTA paradigm.
This work inherits the success of this line, and adopts the pivoting-based scheme as a backbone, but we further strengthen it by leveraging the semantic and syntactic structure information to better solve the two issues.

Scene graphs depict the intrinsic semantic scene structures of images or texts \cite{KrishnaZGJHKCKL17,wang-etal-2018-scene}.
In SGs, the key object and attribute nodes are connected to describe the semantic contexts, which have been shown useful as auxiliary features for wide ranges of downstream applications, e.g., image retrieval \cite{JohnsonKSLSBL15}, image generation \cite{JohnsonGF18} and image captioning \cite{YangTZC19}.
Here we incorporate both the visual and language scene graphs to enhance the cross-modal alignment learning.

Note that \citet{GaoZYJG22} also leverage the SG features for cross-lingual captioning, while ours differs from theirs in three aspects.
First, they consider a fully unsupervised cross-lingual setup with no image-caption pairs at pivot language, while under such an unpaired assumption the visual and language scene graphs are hard to align, and thus limits the utility of SGs.
Second, in this work we sufficiently align the two cross-modal SGs via unsupervised learning, such that the noises in SGs will be effectively screened.
Third, \citet{GaoZYJG22} align the pivot and target languages with also the SG structure.
We note that it could be ineffective to perform cross-lingual alignment based on textual SGs because the scene structures in different languages are essentially the same.

In fact, two languages can be different the most in linguistic structures.
Almost all the erroneous sentences come with certain grammar or syntax errors \cite{jamshid-lou-etal-2019-neural,jamshid-lou-johnson-2020-improving}.
Also syntax features have been extensively found to be effective in improving the language quality (e.g., fluency and grammatically-correctness) in cross-lingual scenario \cite{Nivre15,li-etal-2021-unsupervised-neural,CSynGEC2022}.
For example, in machine translation, different languages show great correspondences in phrasal constituent structures \cite{zhang-zong-2013-learning,fang-feng-2022-neural}.
Also, syntactic structure features have been integrated into a broad number of downstream applications \cite{Wu0RJL21,FeiGraphSynAAAI21,FeiLasuieNIPS22}.
Thus we consider making use of the syntax structures as cross-lingual supervision to enhance the captioning quality.

\vspace{-1mm}
\section{Syntactic Semantic Structure-guided Cross-lingual Captioning Framework}

\vspace{-1mm}
The original task is to learn a mapping $\mathcal{F}_{I\to S^t}$ from input images $I$ to target-language captions $S^t$.
Following \citet{GuJCW18,0003CZJ19}, we decompose $\mathcal{F}_{I\to S^t}$ into two mappings: 1) the image-to-pivot captioning $\mathcal{F}_{I\to S^p}$ training with the paired data $\{(I,S^p)\}$, and 2) the pivot-to-target translation $\mathcal{F}_{S^p\to S^t}$ training with the parallel data $\{(S^p,S^t)\}$.
Note that $\{(I,S^p)\}$ and $\{(S^p,S^t)\}$ are two distinct datasets with possibly no intersection.

In our setting, we also leverage the SG and SC structure features in two mappings.
As shown in Fig. \ref{framework}, 
the semantic structure-guided captioning phase ($\mathcal{F}_{<I,\text{SG}>\to S^p}$) takes as input the image $I$ and the visual SG encoded by a structure encoder, yielding the pivot caption $S^p$.
Then, the syntactic structure-guided translating phase ($\mathcal{F}_{<S^p,\text{SC}>\to S^t}$) takes as input the $S^p$ and the pivot SC,
finally producing the target caption $S^t$.
Note that the input embeddings of the second step are shared with the output embeddings from the first step so as to avoid the isolation of the two parts.
Also we impose a residual connection from the SG feature representations to the SC feature representations to supervise the final target captioning with scene features.


\vspace{-1mm}
\subsection{Semantic Structure-guided Captioning}
Given an image, we obtain its SG from an off-the-shelf SG parser, which is detailed in the experiment setup.
We denote an SG as $SG$=($V,E$), where $V$ is the set of nodes $v_i \in V$ (including object, attribute and relation types),\footnote{
Appendix $\S$\ref{Scene Graph Specification} details the SG and SC structures.
} $E$ is the set of edges $e_{i,j}$ between any pair of nodes $v_i$.
We encode a SG with a graph convolution network \citep[GCN;][]{marcheggiani-titov-2017-encoding}:
\begin{equation}\small\label{GCN-SG}
\setlength\abovedisplayskip{3pt}
\setlength\belowdisplayskip{3pt}
 \{ \bm{h}_{i} \} = \text{GCN}^{G}( SG ) \,,
\end{equation}
where $\bm{h}_{i}$ is the representation of a node $v_i$.
We then use a Transformer \cite{VaswaniSPUJGKP17} decoder to predict the pivot caption $\hat{S}^p$ based on $\{\bm{h}_{i}\}$:
\begin{equation}\small\label{dec-pivot}
\setlength\abovedisplayskip{3pt}
\setlength\belowdisplayskip{3pt}
 \hat{S}^p = \text{Trm}^{G}( \{ \bm{h}_{i} \} ) \,.
\end{equation}

\vspace{-1mm}
\subsection{Syntactic Structure-guided Translation}
\label{Syntactic Structure-guided Translation}

In this step we first transform the predicted pivot caption $S^p$ into the SC structure, $SC$=($V, E$), where $V$ are the phrasal\&word nodes connected by the compositional edge $E$.
Different from the dependency-like SG structure, SC is a tree-like hierarchical structure, as depicted in Fig. \ref{intro}.
Similarly, we encode SC trees with another GCN:
\begin{equation}\small\label{GCN-SC}
\setlength\abovedisplayskip{3pt}
\setlength\belowdisplayskip{3pt}
 \{ \bm{r}_{j} \} = \text{GCN}^{C}( SC ) \,,
\end{equation}
where $\bm{r}_{j}$ is an SC node representation.
Another Transformer decoder is used to predict the target caption $\hat{S}^t$.
To ensure the relevancy of target-side generation, we create a shortcut between the prior SG feature representations $\bm{h}$ and the SC features $\bm{r}$, via the cross-attention mechanism:
\begin{equation}\small\label{dec-target}
\setlength\abovedisplayskip{3pt}
\setlength\belowdisplayskip{3pt}
 \hat{S}^t = \text{Trm}^{C}( \{\bm{r}_{j}\}; \{\bm{h}_{i}\} ) \,.
\end{equation}

\begin{figure}[!t]
\centering
\includegraphics[width=1\columnwidth]{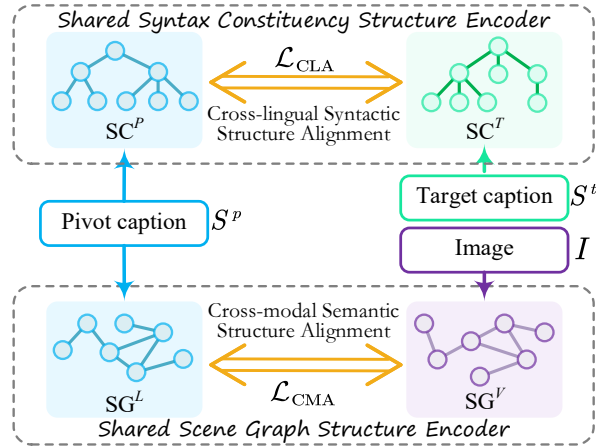}
\caption{
Illustration of the cross-modal semantic and cross-lingual syntactic structure alignment learning.
}
\label{cross-learning}
\end{figure}

\begin{figure*}[!t]
\centering
\includegraphics[width=1\textwidth]{illus/bt-2.pdf}
\caption{
Illustrations of the cross-modal and cross-lingual back-translation.
}
\label{back-trainslation}
\end{figure*}

\vspace{-1mm}
\subsection{Two Separate Supervised Learning}

The captioning and the translation steps are performed separately based on $\{(I,S^p)\}$ and $\{(S^p,S^t)\}$ in a supervised manner:
\setlength\abovedisplayskip{2pt}
\setlength\belowdisplayskip{2pt}
\begin{equation}\small\label{ce-cap}
\mathcal{L}_{\text{\scriptsize Cap}}  = - \sum \log P({S^p} | I,\text{SG} )  \,,
\end{equation}
\begin{equation}\small\label{ce-trans}
\mathcal{L}_{\text{\scriptsize Tran}}  = - \sum \log P({S^t} | S^p,\text{SC} )  \,.
\end{equation}

\vspace{-1mm}
\section{Structure-Pivoting Cross-lingual Cross-modal Alignment Learning}

In the above supervised training, though leveraging the semantic and syntactic structure information, the cross-modal image-text pair and the cross-lingual pivot-target pair can be still under-aligned in their own feature spaces, due to the intrinsic structural gaps, e.g., noisy substructures.
To combat that, we further propose two structure-pivoting unsupervised learning strategies (cf. Fig. \ref{cross-learning}): cross-modal semantic structure alignment and cross-lingual syntactic structure alignment.
Besides, the two parts of our backbone captioner are initially trained separately.
This motivates us to further align the two procedures in a whole-scale way, with cross-lingual\&cross-modal back-translation training (cf. Fig. \ref{back-trainslation}).

\vspace{-1mm}
\subsection{Cross-modal Semantic Structure Aligning}

The basic idea is to encourage those text nodes and visual nodes that serve a similar role in the visual SG$^V$ and language SG$^L$ to be closer, while for those not we hope to repel them from each other, so as to mitigate the scene inconsistency.
We realize this via the current popular CL technique.
We first obtain the node representations of visual SG ($\bm{h}_i^{V}$) and language SG ($\bm{h}_j^{L}$) using one shared GCN encoder as in Eq. (\ref{GCN-SG}), based on the ground-truth $\{(I,S^p)\}$ data.
We then measure the similarities between all pairs of nodes from two SGs:
\setlength\abovedisplayskip{2pt}
\setlength\belowdisplayskip{2pt}
\begin{equation}\small\label{similar-scoring-1}
s^m_{i,j} = \frac{ (\bm{h}^{V}_i)^T \cdot \bm{h}^{L}_j }{ || \bm{h}^{V}_i ||  \, || \bm{h}^{L}_j ||  }  \,.
\end{equation}
A pre-defined threshold $\rho_m$ will decide the alignment confidence, i.e., pairs with $s^m_{i,j}>\rho_m$ is considered similar. Then we have:
\setlength\abovedisplayskip{2pt}
\setlength\belowdisplayskip{2pt}
\begin{equation}\small\label{CL-m-1}
\mathcal{L}_{\text{\scriptsize CMA}}  = - \sum_{i\in \text{SG}^{V} ,\, j^{\ast}\in \text{SG}^{L}} \log 
\frac{\exp(s^m_{i,j^{\ast}} /\tau_m )}{\mathcal{Z}}  \,,
\end{equation}
where $\tau_m$>0 is an annealing factor.
$j^{\ast}$ represents a positive pair with $i$, i.e., $s^m_{i,j^{\ast}}$>$\rho_m$.
$\mathcal{Z}$ is a normalization factor for probability calculation.

\vspace{-1mm}
\subsection{Cross-lingual Syntactic Structure Aligning}

The idea is similar to the above one, while in the cross-lingual syntactic structure space.
We use the shared SC GCN encoder to generate node representations $\bm{r}^P_i$ and $\bm{r}^T_j$ of pivot-/target-side SCs on the parallel sentences.
CL loss is then put on the similarity score $s^l_{i,j}$ to carry out the unsupervised alignment learning, which we summarize as $\mathcal{L}_{\text{\scriptsize CLA}}$.

\vspace{-1mm}
\subsection{Cross-modal\&lingual Back-translation}

Drawing inspiration from unsupervised machine translation, we leverage the back-translation technique \cite{sennrich-etal-2016-improving,edunov-etal-2018-understanding} to achieve the two-step alignment over the overall framework.
We present the cross-lingual cross-modal back-translation training, including the image-to-pivot back-translation and the pivot-to-target back-translation.

\vspace{-1mm}
\paragraph{Image-to-Pivot Back-translation}
With gold image-caption pairs at hand, we can first obtain the target caption prediction $\hat{S}^t$ via our cross-lingual captioner.
We then translate the $\hat{S}^t$ into pseudo pivot caption $\hat{S}^p$ via an external translator $\mathcal{M}_{t\to p}$.
This thus forms a path: $S^p$-$I$$\to$$\hat{S}^t$$\to$$\hat{S}^p$.
And our framework can be updated via:
\setlength\abovedisplayskip{2pt}
\setlength\belowdisplayskip{2pt}
\begin{equation}  \small
\mathcal{L}_{\text{\scriptsize IPB}} = \mathbb{E} [-\log p (\hat{S}^p | \mathcal{M}_{t\to p} (\mathcal{F}_{I\to S^t}(I)) ) ] \,.
\end{equation}

\vspace{-1mm}
\paragraph{Pivot-to-Target Back-translation}
There is a similar story for the gold pivot-target parallel sentences: $S^t$-$S^p$$\to$$\hat{I}$$\to$$\hat{S}^t$.
For $S^p$$\to$$\hat{I}$ we leverage an external SG-based image generator \cite{JohnsonGF18,ZhaoWCG22}.
The learning loss is:
\setlength\abovedisplayskip{2pt}
\setlength\belowdisplayskip{2pt}
\begin{equation} \small
\mathcal{L}_{\text{\scriptsize PTB}} = \mathbb{E} [-\log p (\hat{S}^t | \mathcal{F}_{I\to S^t} (\mathcal{M}_{S^p\to I}(S^p)) ) ] \,.
\end{equation}

\vspace{-1mm}
\paragraph{$\bigstar$ Remarks on Training}
We take a warm-start strategy to ensure stable training of our framework.
Initially we pre-train two parts separately via $\mathcal{L}_{\text{\scriptsize Cap}}$\&$\mathcal{L}_{\text{\scriptsize Trans}}$
We then perform two structure-pivoting unsupervised alignment learning via $\mathcal{L}_{\text{\scriptsize CMA}}$\&$\mathcal{L}_{\text{\scriptsize CLA}}$.
Finally, we train the overall model via back-translation $\mathcal{L}_{\text{\scriptsize IPB}}$\&$\mathcal{L}_{\text{\scriptsize PTB}}$.
Once the system tends to converge, we put them all together for further overall fine-tuning:
\begin{equation}\small
\mathcal{L} = \mathcal{L}_{\text{\scriptsize Cap}} + \mathcal{L}_{\text{\scriptsize Trans}} + \mathcal{L}_{\text{\scriptsize CMA}} + \mathcal{L}_{\text{\scriptsize CLA}}  + \mathcal{L}_{\text{\scriptsize IPB}} + \mathcal{L}_{\text{\scriptsize PTB}} \,.
\end{equation}
Here for brevity, we omit the item weights.
Appendix $\S$\ref{Extended Learning Processing} gives more training details.

\section{Experimental Setups}
\label{Experimental Setups}

\paragraph{Datasets}
To align with existing work, we consider the transfer between English (En) and Chinese (Zh), and use the image caption datasets of MSCOCO \cite{LinMBHPRDZ14}, AIC-ICC \cite{AIC-ICC} and COCO-CN \cite{LiXWLJYX19}.
We use the training set of a language as image-pivot pairs for the first part training, and test with the set of another language.
For the second part training, we collect the paired En-Zh parallel sentences from existing MT data, including UM \cite{TianWCQOY14} and WMT19 \cite{barrault-etal-2019-findings}.

\begin{table*}[!t]
  \centering
\fontsize{9}{10.5}\selectfont
\setlength{\tabcolsep}{2.1mm}
\resizebox{1\textwidth}{!}{
\begin{tabular}{lccccccccc}
\hline
\multicolumn{1}{c}{\multirow{2}{*}{\textbf{}}}& \multicolumn{4}{c}{\textbf{Zh $\to$ En}} & \multicolumn{4}{c}{\textbf{En $\to$ Zh}} & \multicolumn{1}{c}{\multirow{2}{*}{\textbf{\emph{Avg.}}}} \\
\cmidrule(r){2-5}\cmidrule(r){6-9}
 & BLEU & METEOR  & ROUGE &  	CIDEr & 	BLEU & 	METEOR &  	ROUGE &  	CIDEr & 	  \\
 
\hline
\multicolumn{10}{l}{$\bullet$ \textbf{\emph{Translation-based methods}}}\\

EarlyTranslation & 	48.3 & 	15.2 & 	27.2 & 	18.7 & 	43.6  & 	20.3 &  	30.3 & 	14.2 & 	27.2 \\
LateTranslation & 	45.8  & 	13.8  & 	25.7 & 	14.5 & 	41.3 & 	13.5 & 	26.7 & 	14.0 & 	24.4 \\
FG & 	46.3 & 	12.5 & 	25.3 & 	15.4 & 	43.0 & 	19.7 & 	29.7 & 	15.7 & 	25.9 \\
SSR$^{\dagger}$ &  	52.0  & 	14.2  & 	27.7 & 	28.2 & 	46.0  & 	22.8  & 	32.0 & 	18.3 & 	30.1 \\
\cdashline{1-10}
\multicolumn{10}{l}{$\bullet$ \textbf{\emph{Pivoting-based methods}}}\\
PivotAlign & 	52.1 & 	17.5 & 	28.3 & 	27.0 & 	47.5 & 	23.7 & 	32.3 & 	19.7 & 	31.1 \\
UNISON & 	54.3 & 	18.7 & 	30.0 & 	28.4 & 	48.7 & 	25.2 & 	33.7 & 	21.9 & 	32.4 \\
\rowcolor{nmgray}  \textsc{Cross$^2$StrA} (Ours) & 	 \bf 57.7 &	 \bf21.7 &	 \bf 33.5 &	\bf 30.7 &	\bf 52.8 &	\bf 27.6 &	\bf 36.1 &	\bf 24.5 &	\bf 35.8 \\
\quad w/o SG & 	55.8 &	19.1 &	31.2 &	28.0 &	48.6 &	25.8 &	33.9 &	21.6 &	33.1
 \\
\quad w/o SC & 	56.1 &	20.0 &	32.1 &	28.9 &	50.4 &	26.6 &	35.4 &	23.3 &	34.1
 \\
\quad w/o ResiConn & 	56.4 &	21.2 &	32.9 &	29.4 &	51.8 &	27.1 &	35.9 &	24.1 &	34.9 \\

    \hline
    \end{tabular}%
    }
    \caption{Transfer results between MSCOCO (En) and AIC-ICC (Zh).
    The values of SSR$^{\dagger}$ are copied from \citet{0003CZJ19}, while all the rest are from our implementations.
    }
  \label{tab:main}%
\end{table*}%

\paragraph{Baselines and Evaluations}
Our comparing systems include 1) the translation-based methods, including the \emph{early translation} and \emph{late translation} mentioned in the introduction, \emph{FG} \cite{LanLD17}, \emph{SSR} \cite{0003CZJ19}, and 2) the pivoting-based methods, including \emph{PivotAlign} \cite{GuJCW18} and \emph{UNISON} \cite{GaoZYJG22}.
Following baselines, we report the BLEU \cite{papineni-etal-2002-bleu}, METEOR \cite{denkowski-lavie-2014-meteor}, ROUGE \cite{lin-2004-rouge} and CIDEr \cite{VedantamZP15} scores for model evaluation.
Our results are computed with a model averaging over 10 latest checkpoints.

\paragraph{Implementations}

To obtain the visual SGs, we employ the FasterRCNN \cite{RenHGS15} as an object detector, and MOTIFS \cite{ZellersYTC18} as a relation classifier and an attribute classifier.
For language SGs, we first convert the sentences into dependency trees with a parser \cite{00010BT0GZ18}, and then transform them into SGs based on certain rules \cite{schuster-etal-2015-generating}.
We obtain the SC trees via the Berkeley Parser \cite{kitaev-klein-2018-constituency}, trained on PTB \cite{MarcusSM94} for En texts and on CTB \cite{XueXCP05} for Zh texts.
In our back-translation learning, we use the T5 \cite{RaffelSRLNMZLL20} as the target-to-pivot translator ($\mathcal{M}_{t\to p}$), and adopt the current SoTA SG-based image generator ($\mathcal{M}_{S^p\to I}$) \cite{ZhaoWCG22}.
Chinese sentences are segmented via Jieba\footnote{\url{https://github.com/fxsjy/jieba}}.
We use Transformer to offer the underlying textual representations for GCN, and use FasterRCNN \cite{RenHGS15} for encoding visual feature representations. 
Our SG and SC GCNs and all other embeddings have the same dimension of 1,024.
All models are trained and evaluated with NVIDIA A100 Tensor Core GPUs.

\vspace{-2mm}
\section{Experimental Results and Analyses}
\label{Experimental Results and Analyses}

\begin{table}[t]
  \centering
\fontsize{9}{11.5}\selectfont
 \setlength{\tabcolsep}{1.4mm}
\resizebox{1\columnwidth}{!}{
\begin{tabular}{lccccl}
\hline

\multicolumn{1}{c}{\multirow{2}{*}{\textbf{}}}& \multicolumn{2}{c}{\textbf{Zh $\to$ En}} & \multicolumn{2}{c}{\textbf{En $\to$ Zh}} & \multicolumn{1}{c}{\multirow{2}{*}{\textbf{\emph{Avg.}}}} \\
\cmidrule(r){2-3}\cmidrule(r){4-5}
&B&R&B&R&\\
 \hline
\rowcolor{nmgray} \textsc{Cross$^2$StrA} & \bf 57.7 &	\bf 33.5 &	\bf 52.8 &	\bf 36.1 &	\bf 45.0 \\
\quad w/o $L_{\text{\scriptsize CMA}}$ &	54.4	&	29.7	&	50.1	&	34.9  &	42.3\scriptsize{(-2.7)} \\
\quad w/o $L_{\text{\scriptsize CLA}}$ &	54.6	&	30.1	&	51.0	&	35.3 &	43.0\scriptsize{(-2.0)} \\
\quad w/o $L_{\text{\scriptsize IPB}}$ &	53.8	&	31.1	&	50.5	&	35.1 &	43.1\scriptsize{(-1.9)} \\
\quad w/o $L_{\text{\scriptsize PTB}}$ &	55.0	&	32.8	&	52.2	&	35.7  &	44.2\scriptsize{(-0.8)} \\
\cdashline{1-6}
\quad w/o $L_{\text{\scriptsize CMA}}$+$L_{\text{\scriptsize CLA}}$ &	51.8	&	27.7	&	47.5	&	33.7 &	40.8\scriptsize{(-4.2)} \\
\quad w/o $L_{\text{\scriptsize IPB}}$+$L_{\text{\scriptsize PTB}}$ &	52.7	&	30.1	&	49.9	&	34.2 &	42.2\scriptsize{(-2.8)} \\
\hline
\end{tabular}
}
\caption{
Ablating different learning strategies.
B: BLEU, R: ROUGE.
}
\label{tab:Ablation}
\end{table}

\vspace{-1mm}
\paragraph{Transfer between MSCOCO and AIC-ICC}

Table \ref{tab:main} presents the Zh$\to$En and En$\to$Zh transfer results.
We first can observe that the \emph{EarlyTranslation} is more effective than \emph{LateTranslation}, as the former introduces lesser noises in training.
Also, we see that among all the translation-based methods, \emph{SSR} shows the best performance.
Further, it is clear that the pivoting methods show overall better results than the translation ones.
This is most possibly because the joint training in pivoting-based models relieves the under-alignment between the captioning and translation stages, reducing the noise introduction of the pipeline.

\begin{table*}[!t]
  \centering
\fontsize{9}{10.5}\selectfont
\setlength{\tabcolsep}{2.3mm}
\resizebox{1\textwidth}{!}{
\begin{tabular}{lcccccccc}
\hline
 & \bf BLEU@1  &	\bf BLEU@2 & \bf BLEU@3 & 	\bf BLEU@4  &	\bf METEOR 	 & \bf ROUGE  &	 \bf CIDEr & \textbf{\emph{Avg.}}	  \\
\hline
\multicolumn{9}{l}{$\bullet$ \textbf{\emph{Translation-based methods}}}\\
EarlyTranslation$^{\dagger}$ &	60.4 & 	40.7  &	26.8  &	17.3  &	24.0 & 	43.6  &	52.7 &	37.9\\
LateTranslation$^{\dagger}$ &	58.9  &	38.0  &	23.5  &	14.3  &	23.5  &	40.2  &	47.3 &	35.1\\
SSR &	65.2 &	43.5 &	27.3 &	17.7 &	25.4 &	45.9 &	53.8 &	39.8\\
\cdashline{1-9}
\multicolumn{9}{l}{$\bullet$ \textbf{\emph{Pivoting-based methods}}}\\
PivotAlign &	66.5 &	45.0 &	29.3 &	18.2 &	27.0 &	46.3 &	55.0 &	41.0\\
UNISON$^{*\dagger}$ &	63.4 & 	43.2  &	29.5  &	17.9  &	24.5  &	45.1  &	53.5 &	39.5\\
UNISON &	68.3 &	46.7 &	30.6 &	19.0 &	29.4 &	48.0 &	56.3	 &42.7\\
\rowcolor{nmgray}  \textsc{Cross$^2$StrA} &	 \bf 70.4 &	 \bf 48.8 &	 \bf 32.5 &	 \bf 20.8 &	 \bf 31.9 &	 \bf 50.6 &	 \bf 58.2  &\bf 44.7\\
    \hline
    \end{tabular}%
    }
    \caption{Cross-lingual transfer from MSCOCO (En) to COCO-CN (Zh).
    The values with ${\dagger}$ are copied from \citet{0003CZJ19}.
    UNISON$^{*\dagger}$ is the raw version without using the paired image-caption(pivot) data for training.
    }
  \label{tab:main2}%
\end{table*}%

Looking into the pivoting-based models, \emph{UNISON} exhibits the stronger capability of the transfer in both directions, owing to the integration of SG structure features, i.e., helping accurately capture the semantic relevances between vision and language.
Most importantly, our \textsc{Cross$^2$StrA} outperforms all the other baselines with significant margins on all metrics consistently.
For example, we improve over \emph{UNISON} by 3.4 (Zh$\to$En) and 4.1 (En$\to$Zh) BLEU scores respectively.
We give credit to the integration of both the semantic SG and the syntactic SC structures, as well as the effective alignment learning strategies.
The above observations show the efficacy of our system for cross-lingual captioning.

\begin{figure}[!t]
\centering
\includegraphics[width=1\columnwidth]{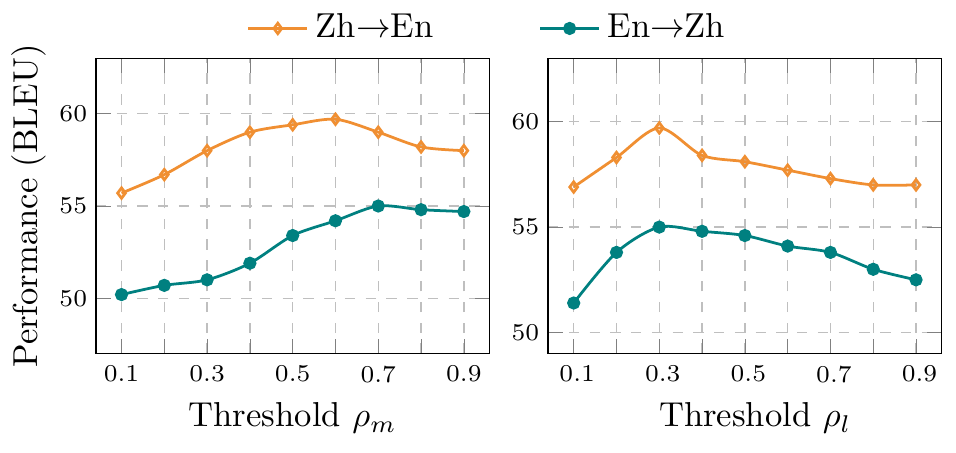}
\caption{
Influences of different threshold values in two structure-guided alignment learning.
}
\label{threshold}
\end{figure}

\vspace{-1mm}
\paragraph{Influences of Learning Strategies}
In Table \ref{tab:Ablation} we quantify the contribution of each learning objective via ablation.
As seen, each learning strategy shows the impact to different extents.
For example, the cross-modal semantic alignment gives greater influences than the cross-lingual syntactic alignment of the overall performances (i.e., 2.7 vs. 2.0).
In contrast to the two structure-pivoting learning ($L_{\text{\scriptsize CMA}}$+$L_{\text{\scriptsize CLA}}$), we can find that the back-translation learning ($L_{\text{\scriptsize IPB}}$+$L_{\text{\scriptsize PTB}}$) shows slightly lower impacts.
Particularly the pivot-to-target back-translation contributes limitedly, and we believe the quality of SG-to-image generator should bear the responsibility.

\vspace{-1mm}
\paragraph{Threshold Study}

In Fig. \ref{threshold} we study the influences of threshold values on the two alignment learning, by varying $\rho_m$ and $\rho_l$.
As seen, when $\rho_m$ is 0.6 and 0.7 in two tasks respectively, the overall transfer results are the best, while $\rho_l$=0.3 helps give the best effects.
Such a pattern distinction between $\rho_m$ and $\rho_l$ implies that the SGs between vision and language have less discrepancy, while the SC structures between two languages come with non-negligible differences.

\vspace{-1mm}
\paragraph{Transfer from MSCOCO to COCO-CN}
Table \ref{tab:main2} further shows the transfer results from English MSCOCO to Chinese COCO-CN.
The overall tendency is quite similar to the one in Table \ref{tab:main}.
We see that translation methods are inferior to the pivoting methods.
Our \textsc{Cross$^2$StrA} model gives the best performances on all metrics, outperforming \emph{UNISON} by an average 2.0(=44.7-42.7) score.
This again verifies the efficacy of our proposed method.

\begin{figure}[!t]
\centering
\includegraphics[width=0.97\columnwidth]{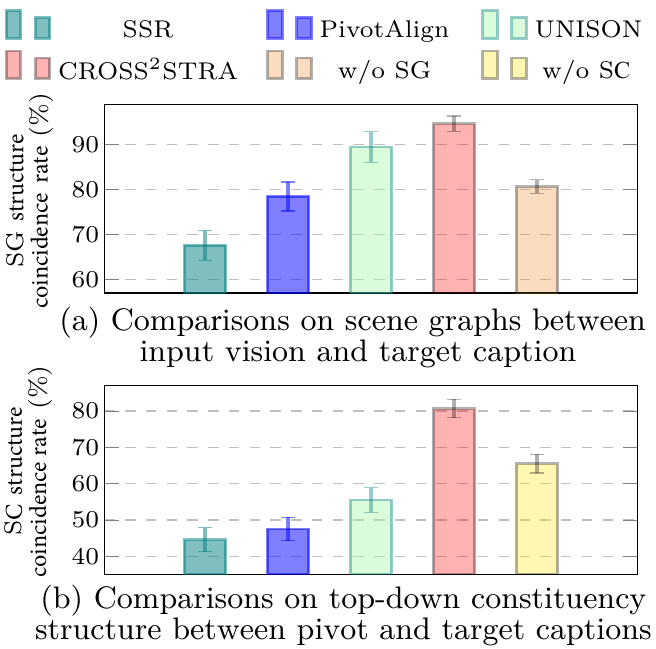}
\caption{
The matchings of SG and SC structures.
}
\label{divergence}
\end{figure}

\begin{figure*}[!t]
\centering
\includegraphics[width=1\textwidth]{illus/case-2.pdf}
\vspace{-1mm}
\caption{
Qualitative results of cross-lingual captioning.
The instances are randomly picked from AIC-ICC (Zh).
}
\label{case-study}
\end{figure*}

\vspace{-1mm}
\paragraph{Probing Cross-modal and Cross-lingual Structure Alignment}
We integrate the semantic scene structure and syntactic structures with the aim of better cross-modal and cross-lingual alignment in our two-stage pivoting transfer framework.
Here we directly assess to what extent our methods improve the alignment.
Fig. \ref{divergence} shows the structure coincidence rate between the input image SG and predicted target caption SG, and the SC structure coincidence rate between the pivot and target captions.\footnote{
Appendix $\S$\ref{Specification on Structure Coincidence Probing} details the measuring method.
}
We see that with the integration of semantic scene modeling, both \emph{UNISON} and our system exhibit prominent cross-modal alignment ability, i.e., with higher structural overlaps.
The same observation can be found with respect to syntactic structure integration for enhancing cross-lingual alignment learning.
Either without the leverage of SG or SC structure, the corresponding cross-modal or cross-lingual alignment effect is clearly weakened.

\begin{table}[!t]
  \centering
\fontsize{9}{11.5}\selectfont
\setlength{\tabcolsep}{0.4mm}
\resizebox{1\columnwidth}{!}{
\begin{tabular}{lccc}
\hline
 & \bf Relevancy$\uparrow$	&\bf Diversification$\uparrow$	& \bf Fluency$\uparrow$ \\
\hline
FG &	5.34 &3.75 &	7.05 \\
SSR	 & 7.86 &5.89 &	7.58 \\
PivotAlign &	8.04 &6.57 &	7.46 \\
UNISON &	9.02 &9.14 &	7.89 \\
\rowcolor{nmgray}  \textsc{Cross$^2$StrA} &	\bf 9.70$^{\ddagger}$ &\bf 9.53$^{\ddagger}$ &	\bf 9.22$^{\ddagger}$ \\
\quad w/o SG &	8.35 &7.75 &	9.04 \\
\quad w/o SC &	9.42 &8.34 &	8.07 \\
\quad w/o $L_{\text{\scriptsize CMA}}$+$L_{\text{\scriptsize CLA}}$ &	7.80 &7.24 &	8.15 \\
\hline
\end{tabular}%
}
\caption{
Human evaluations are rated on a Likert 10-scale.
$\ddagger$ indicates significant better over the baselines ($p$<0.03).
}
  \label{human}%
\end{table}%

\vspace{-1mm}
\paragraph{Human Evaluation}

We further try to quantify the improvements of the generated captions via human evaluation.
In Table \ref{human} we show the evaluation results based on MSCOCO (En) to AIC-ICC (Zh) transfer, on three dimensions: \emph{relevancy}, \emph{diversification} and \emph{fluency}.
We can see that our system shows significantly higher scores than baseline systems in terms of all three indicators.
For those methods with SG structure features, the content relevancy and diversification of captions are much better.
Yet only our method gives satisfied language fluency, due to the equipment of syntactic features.
With further ablation studies we can further confirm the contributions of the SG and SC features.

\begin{figure}[!t]
\centering
\includegraphics[width=0.98\columnwidth]{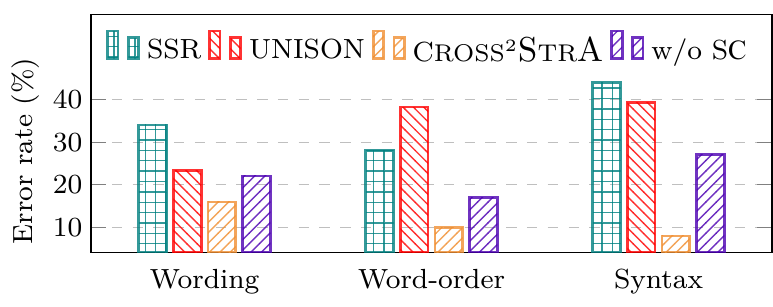}
\caption{
Evaluating the language quality of captions.
}
\label{linguistic-quality}
\end{figure}

\vspace{-1mm}
\paragraph{Captioning Linguistic Quality Study}
We take a further step, investigating how exactly our model improves the linguistic quality of the target captions.
Same to the human evaluation, we ask native speakers to measure the errors that occurred in the generated captions, in terms of \emph{wording}, \emph{word order} and \emph{syntax correctness}.
Fig. \ref{linguistic-quality} presents the results of the transfer from MSCOCO (En) to AIC-ICC (Zh).
We see that our model has committed the least errors, where the performances on syntax correctness are especially higher than baselines.
Once without using the syntactic features, the error rates grow rapidly, which demonstrates the importance to integrate the syntactic structures.

\vspace{-1mm}
\paragraph{Qualitative Result}
Finally, we empirically show some real prediction cases, so as to aid an intuitive understanding of our method's strength.
In Fig. \ref{case-study} we provide four pieces of testing examples on the En$\to$Zh transfer, which we compare with different baseline methods.
As can be seen, the \emph{SSR} model often tends to generate target-side captions with lower diversification, and meanwhile unsatisfactory content relevancy, and thus inaccurate image descriptions.
On the contrary, the captions from \emph{UNISON} are much better, i.e., better relevancy and diversification.
We can give credit to the equipment of scene graph-based alignment learning.
However, \emph{UNISON} can fall short on language quality, i.e., problematic fluency.
Since English and Chinese differ much in linguistic and grammar characteristics, without leveraging the syntactic structure features, it leads to inferior language quality.
Luckily, our model can address all those issues, and generate captions with good relevancy, diversification, and fluency.
This again proves the effectiveness of our proposed method.

\vspace{-1mm}
\section{Conclusion and Future Work}

\vspace{-1mm}
In this paper we investigate the incorporation of semantic scene graphs and syntactic constituency structure information for cross-lingual image captioning.
The framework includes two phrases, semantic structure-guided image-to-pivot captioning and syntactic structure-guided pivot-to-target translating.
We take the SG and SC structures as pivots, performing cross-modal semantic structure alignment and cross-lingual syntactic structure alignment learning.
A cross-lingual\&cross-modal back-translation training is further performed to align two phrases.
On English$\leftrightarrow$Chinese transfer experiments, our model shows great superiority in terms of captioning relevancy and fluency.

Bridging the gaps between the cross-modal and cross-lingual transfer with external semantic and syntactic structures has shown great potential.
Thus it is promising to extend the idea to other scenarios.
Also, exploiting the external structures potentially will introduce noises,
and thus a dynamical structure induction is favorable.

\section*{Limitations}
\vspace{-2mm}

In this work, we take the sufficient advantages of the external semantic and syntactic structure knowledge to improve our focused problem.
But this could be a double-edged sword to use such features.
Specifically, our paper has the following two potential limitations.
First of all, our method closely relies on the availability of the resources of scene graph structures and syntax structures.
While most of the languages come with these structure annotations to train good-performing structure parsers (for example, the syntax structure annotations of Penn TreeBank cover most of the existing languages), some minor languages may not have structure resources.
That being said, our idea still works well even in the absence of the target-side structure annotations.
With only the structure annotations at pivot-side (resource-rich) language (in this case, the cross-modal semantic\&syntactic structure aligning learning are canceled), we can still achieve much better performances than those baselines without using the structural features.
Besides, our method will be subject to the quality of the external structure parsers.
When the parsed structures of scene graphs and syntax trees are with much noise, the helpfulness of our methods will be hurt.
Fortunately, the existing external semantic and syntactic structure parsers have already achieved satisfactory performances, which can meet our demands.



\bibliography{anthology}
\bibliographystyle{acl_natbib}

\begin{figure*}[!t]
\centering
\includegraphics[width=.96\textwidth]{illus/appendix-SC.pdf}
\caption{
Illustration of the full constituency syntax structure.
}
\label{appendix-SC}
\vspace{-4mm}
\end{figure*}

\newpage

\appendix

\section{Model Details}
\label{Model Details}

\subsection{Specification of Scene Graph and Syntax Constituency Structures}
\label{Scene Graph Specification}

In Fig. \ref{appendix-SC} and Fig. \ref{appendix-SG} we illustrate the complete structures of the syntactic constituency tree and scene graphs, respectively.
We note that the scene graph is a dependency-like structure, describing the node-node inter-relation in an `is-a' paradigm.
And the syntactic constituency tree is a compositional structure, depicting how words constitute phrases and then organize them into whole sentences.

\begin{figure}[!t]
\centering
\includegraphics[width=1\columnwidth]{illus/appendix-SG2.pdf}
\caption{
Illustration of the full scene graph structures.
The object nodes are the real entities of visual proposals or textual words/phrases.
The attribute nodes and relation nodes are the descriptive words, which, in VSG are automatically generated, and in LSG are extracted from sentences.
}
\label{appendix-SG}
\vspace{-4mm}
\end{figure}

\subsection{Pivot-to-target Translation}

In Eq. (\ref{dec-target}) we use a Transformer decoder to predict the target caption $\hat{S}^t$.
A cross-attention mechanism is first used to fuse the prior SG feature representations $\bm{h}$ and the SC features $\bm{r}$.
Specifically, 
\begin{equation}\nonumber
\setlength\abovedisplayskip{2pt}
\setlength\belowdisplayskip{2pt}
 \bm{e} = \text{Softmax}( \frac{ \bm{r} \oplus \bm{h} }{ \sqrt{d} }) \cdot \bm{r}  \,,
\end{equation}
where $d$ is a scaling factor.
Then, the Transformer performs decoding over $\{\bm{e}\}$:
\begin{equation}\nonumber
\setlength\abovedisplayskip{3pt}
\setlength\belowdisplayskip{3pt}
 \hat{S}^t = \text{Trm}^{C}( \{\bm{e}\} ) \,.
\end{equation}

\subsection{Specification on Contrastive Learning}
\label{Specification on Contrastive Learning}

\paragraph{Cross-modal Semantic Structure Aligning}
In Eq. (\ref{CL-m-1}) we define the contrastive learning objective of cross-modal semantic structure aligning, here we unfold the equation:
\begin{equation}\nonumber
\mathcal{L}_{\text{\scriptsize CMA}}  = - \sum_{i\in \text{SG}^{V} ,\, j^{\ast}\in \text{SG}^{L}} \log 
\frac{\exp(s^m_{i,j^{\ast}} /\tau_m )}{\mathcal{Z}}  \,,
\end{equation}
\begin{equation}\nonumber
\mathcal{Z} = \sum_{i\in \text{SG}^{V} ,\, k\in \text{SG}^{L},\, k \ne j^{\ast}}\exp(s_{i,k}/\tau_m)  \,,
\end{equation}
where $\tau_m$>0 is an annealing factor.
$j^{\ast}$ represents a positive pair with $i$, i.e., $s^m_{i,j^{\ast}}$>$\rho_m$.

\paragraph{Cross-lingual Syntactic Structure Aligning}

We detail the cross-lingual syntactic structure aligning learning objective here: 
\begin{equation}\nonumber
\mathcal{L}_{\text{\scriptsize CMA}}  = - \sum_{i\in \text{SC}^{P} ,\, j^{\ast}\in \text{SC}^{T}} \log 
\frac{\exp(s^l_{i,j^{\ast}} /\tau_l )}{\mathcal{Z}}  \,,
\end{equation}
\begin{equation}\nonumber
\mathcal{Z} = \sum_{i\in \text{SC}^{P} ,\, k\in \text{SC}^{T},\, k \ne j^{\ast}}\exp(s_{i,k}/\tau_l)  \,,
\end{equation}
where $\tau_l$>0 is an annealing factor.
$j^{\ast}$ represents a positive pair with $i$, i.e., $s^m_{i,j^{\ast}}$>$\rho_m$.

\subsection{Specifying Overall Training Processing}
\label{Extended Learning Processing}

The training of our framework is based on the warm-up strategy, including four stages.

\textbf{At the first stage,} we use the paired image-caption data  $\{(I,S^p)\}$ at the pivot language side (as well as the VSG structure features) to train the captioning part of our model; and use the parallel sentences $\{(S^p,S^t)\}$ (as well as the pivot-side syntax tree features) to train the translation part of our model, both of two training is supervised.

\textbf{At the second stage,} we perform two structure-pivoting unsupervised alignment learning, by using the image-caption data  $\{(I,S^p)\}$, parallel sentences $\{(S^p,S^t)\}$, and the two structure resource.

\textbf{At the third stage,} we perform the cross-modal and cross-lingual back-translation learning.
This is a whole-framework-level training, aiming to sufficiently align the captioning and translation parts.

\textbf{At the fourth stage,} the system tends to converge, and we put all the above learning objects together for further overall fine-tuning:
\begin{equation}
\begin{aligned} \nonumber
\mathcal{L} &= \lambda_{\text{\scriptsize Cap}} \mathcal{L}_{\text{\scriptsize Cap}} + \lambda_{\text{\scriptsize Trans}} \mathcal{L}_{\text{\scriptsize Trans}}   \, \\
&+ \lambda_{\text{\scriptsize CMA}} \mathcal{L}_{\text{\scriptsize CMA}} + \lambda_{\text{\scriptsize CLA}} \mathcal{L}_{\text{\scriptsize CLA}}     \, \\
&+ \lambda_{\text{\scriptsize IPB}} \mathcal{L}_{\text{\scriptsize IPB}}
+ \lambda_{\text{\scriptsize PTB}} \mathcal{L}_{\text{\scriptsize PTB}} \,.
\end{aligned}
\end{equation}
Here $\lambda_{*}$ are the loss weights that dynamically change by linearly learning scheduler \cite{huang-etal-2020}.
The initial weights are given as: $\lambda_{\text{\scriptsize Cap}}$=1, $\lambda_{\text{\scriptsize Trans}}$=1,
$\lambda_{\text{\scriptsize CMA}}$=0.7, $\lambda_{\text{\scriptsize CLA}}$=0.7, 
$\lambda_{\text{\scriptsize VCB}}$=0.3, $\lambda_{\text{\scriptsize CPB}}$=0.3.
$\lambda_{\text{\scriptsize Cap}}$ and $\lambda_{\text{\scriptsize Trans}}$ will be linearly decreased from 1 to 0.7 along the training,
$\lambda_{\text{\scriptsize CMA}}$ and $\lambda_{\text{\scriptsize REC}}$ are kept unchanged, while $\lambda_{\text{\scriptsize VCB}}$ and $\lambda_{\text{\scriptsize CPB}}$ will be decreased from 0.3 to 0.7.

\section{Extended Experiment Setups}
\label{Extended Experiment Setups}

\subsection{Dataset Description}

We use three image captioning datasets $\{(I,S^p)\}$: MSCOCO, AIC-ICC and COCO-CN.
All the data split follows the same practice as in prior cross-lingual image captioning works \cite{AIC-ICC,0003CZJ19}.
The MSCOCO dataset is annotated in English, which consists of 123,287 images and 5 manually labeled English captions for each image. 
We utilize 113,287 images for training, 5,000 images for validation, and 5,000 images for testing.
The AIC-ICC dataset contains 238,354 images and 5 manually annotated Chinese captions for each image. 
There are 208,354 and 30,000 images in the official training and validation set. 
Since the annotations of the testing set are unavailable in the AIC-ICC dataset, we randomly sample 5,000 images from its validation set as our testing set.
The COCO-CN dataset contains 20,342 images and 27,218 caption texts in Chinese.
We use 18,342 images for training, 1,000 for development, and 1,000 for testing.
Table \ref{dataset} gives the detailed statistics of the image captioning data.

For the translation data $\{(S^p,S^t)\}$, we collect about 1M of raw paired En-Zh parallel sentences from the UM \cite{TianWCQOY14} and WMT19 \cite{barrault-etal-2019-findings} machine translation corpus.
We filter the sentences in MT datasets according to an existing caption-style dictionary and resulting in a total of 400,000 parallel sentences.
For the translation training, we use 390,000 sentence pairs for training, 5,000 sentence pairs for validation, and 5,000 pairs for testing.

\begin{table}[!t]
  \centering
\fontsize{9}{10.5}\selectfont
\setlength{\tabcolsep}{2.mm}
\resizebox{1\columnwidth}{!}{
\begin{tabular}{lclrr}
\hline
\multicolumn{1}{l}{\multirow{1}{*}{Dataset}}& \multicolumn{1}{c}{\textbf{Lang.}}&\textbf{Split}& \multicolumn{1}{c}{\textbf{\#Image}} & \multicolumn{1}{c}{\textbf{\#Caption}} \\
\hline
\multirow{4}{*}{MSCOCO}& \multirow{4}{*}{En} &Total& 123,287 & 616,435 \\
& & Train&  113,287 & 566,435 \\
& & Develop&  5,000 & 25,000 \\
& & Test&  5,000 & 25,000 \\
\cdashline{1-5}
\multirow{4}{*}{AIC-ICC}& \multirow{4}{*}{Zh} &Total& 238,354 & 1,191,770 \\
& & Train&  208,354 & 1,041,770 \\
& & Develop&  25,000 & 125,000 \\
& & Test&  5,000 & 25,000 \\
\cdashline{1-5}
\multirow{4}{*}{COCO-CN}& \multirow{4}{*}{Zh} &Total& 20,342 & 27,218 \\
& & Train&  18,342 & 25,218 \\
& & Develop&  1,000 & 1,000 \\
& & Test&  1,000 & 1,000 \\

\hline
\end{tabular}%
}
\caption{
Statistics of image captioning datasets.
}
\vspace{-4mm}
  \label{dataset}%
\end{table}%

\subsection{Specification on Structure Coincidence Probing}
\label{Specification on Structure Coincidence Probing}

In Fig. \ref{divergence} we assess the ability of our model on the cross-modal and cross-lingual structure alignment, by measuring the structure coincidence between the gold one and the one learned by our model.
Here we detail the evaluation setup.

For the semantic scene structures, we evaluate the coincidence between the input images' SGs and the SGs of predicted target-side captions.
These SG structures are parsed by the same methods introduced above.
We then make statistics of the overlapped node pairs between the two SGs as the coincidence rate $\beta^{G}$.
\begin{equation}\nonumber
\beta^{G} = \frac{\text{SG}^V \cap \text{SG}^L }{\text{SG}^V \cup \text{SG}^L } \,,
\end{equation}
where $\text{SG}^V$ and $\text{SG}^L$ denote any word-pair structure of visual SG and target language SG, respectively.

For the syntax structures, we evaluate the coincidence rate of the constituency tree structures between the intermediate pivot captions and the final predicted target-side captions.
(Because the input images come without the syntax trees.)
The SC structures of two languages are parsed using the parsers introduced above.
We note that the divergences of syntax between two languages can be much larger, compared with the divergences of semantic scene structures.
Different from the measurement for SG structure to traverse the whole graph equally, we measure the SC structure coincidence rate  $\beta^{C}$ in a top-down manner.
Specifically, we traverse the constituency trees in a top-down order, and those matched phrasal nodes at a higher level (lower traversing depth from the root node) will have higher scores than those at a lower level.
\begin{equation}\nonumber
\beta^{C} = \frac{(\text{SC}^P \cap \text{SC}^T) \cdot d}{\text{SC}^P \cup \text{SC}^T } \,,
\end{equation}
where $\text{SC}^P$ and $\text{SC}^T$ denote the phrasal constituent structures of the pivot and target language, respectively.
$d$ is a weight, which is defined as the reciprocal of a top-down traversing depth.

\subsection{Specifications of Human Evaluation Standards}
Table \ref{human} shows the human evaluation results.
Specifically, we design a Likert 10-scale to measure the relevancy, diversification, and fluency of the generated target-side captions.
The 10-scale metrics are defined as: 
1-Can't be worse,
2-Terrible,
3-Poor, 
4-Little poor, 
5-Average,
6-Better than average,
7-Adequate, 
8-Good, 
9-Very good,
10-Excellent. 
We ask ten native Chinese speakers to score the results.
And for each result, we use the averaged scores.

In Fig. \ref{linguistic-quality} we also measure the language quality of captions in terms of \emph{wording}, \emph{word order}, and \emph{syntax correctness}.
We ask the same ten native Chinese speakers to score the error degree of these metrics, each of which is defined as:
\begin{itemize}
    \item \textbf{Wording}: Is the choice of words in the captions suitable and precise to describe the input images?
    \item \textbf{Word order}: Are the words, phrases, and components organized correctly and properly in captioning sentences?
    \item \textbf{Syntax correctness}: Are there syntactic errors in the caption texts? such as omitting or repeating words, mixing up verb tenses or verb conjugations, missing prepositions, etc.
\end{itemize}

\end{document}